\documentclass[runningheads]{llncs}
\usepackage{graphicx}
\usepackage{amsmath,amssymb} 
\usepackage{color}
\usepackage[width=122mm,left=12mm,paperwidth=146mm,height=193mm,top=12mm,paperheight=217mm]{geometry}
\usepackage{algorithm2e}
\usepackage{subfig}
\usepackage{bbm}
\begin{document}
\pagestyle{headings}
\mainmatter

\title{Adversarial Training For Sketch Retrieval} 

\titlerunning{Adversarial Training For Sketch Retrieval}

\authorrunning{Antonia Creswell \& Anil Anthony Bharath}

\author{Antonia Creswell \& Anil Anthony Bharath}


\institute{BICV Group, Bioengineering,\\
	Imperial College London\\
	\email{ac2211@ic.ac.uk}
}

\maketitle

\begin{abstract}
Generative Adversarial Networks (GAN) are able to learn excellent representations for unlabelled data which can be  applied to image generation and scene classification. Representations learned by GANs have not yet been applied to retrieval. In this paper, we show that the representations learned by GANs can indeed be used for retrieval. We consider heritage documents that contain unlabelled Merchant Marks, sketch-like symbols that are similar to hieroglyphs. We introduce a novel GAN architecture with design features that make it suitable for sketch retrieval. The performance of this sketch-GAN is compared to a modified version of the original GAN architecture with respect to simple invariance properties. Experiments suggest that sketch-GANs learn representations that are suitable for retrieval and which also have increased stability to rotation, scale and translation compared to the standard GAN architecture.
\end{abstract}

\section{Introduction}
Recently, the UK's National Archives has collected over $70,000$ heritage documents that originate between the $16^{th}$ and $19^{th}$ centuries. These documents make up a small part of the ``Prize Papers", which are of gross historical importance, as they were used to establish legitimacy of ship captures at sea.

\begin{figure}
    \centering
    \includegraphics[width=\textwidth]{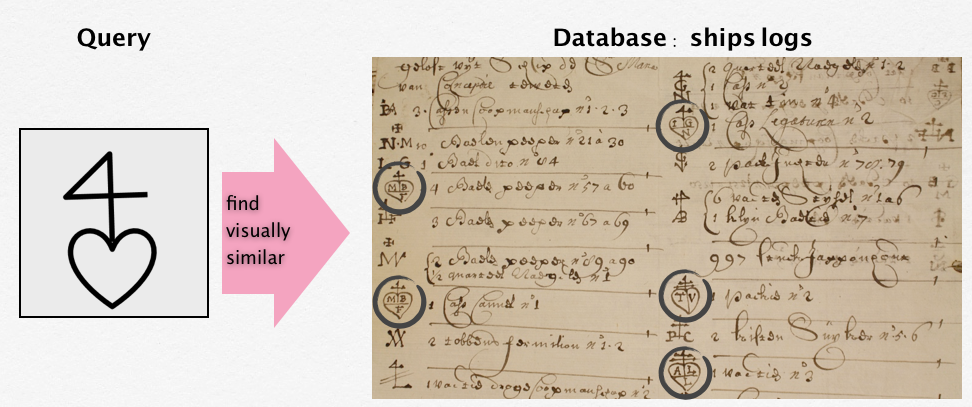}
    \caption{An overview of the problem: the circled items contain examples of Merchant Marks; note that although some marks are distinct, they are still visually similar. We would like to retrieve visually similar examples, and find exact matches if they exist. Note that the two marks on the left are exact matches, while the others might be considered to be visually similar.}
    \label{problem}
\end{figure}

This collection of documents contain \textit{Merchant Marks} (see Fig.\ref{overview}B), symbols used to uniquely identify the property of a merchant. For further historical research to be conducted, the organisation requires that the dataset be searchable by visual example (see Fig.\ref{problem}). These marks are sparse line drawings, which makes it challenging to search for visually similar Merchant Marks between documents. This dataset poses the following challenges to  learning representations that are suitable for visual search:
\begin{enumerate}
    \item Merchant marks are line drawings, absent of both texture and colour, which means that marks cannot be distinguished based on these properties.
    \item Many machine learning techniques, and most notably convolutional neural networks (CNNs), require large amounts of labelled training data, containing on the order of millions of labelled images \cite{krizhevsky2012imagenet}. None of the Merchant Marks are labelled, and in many cases it is not clear what labels would be assigned to them. This motivates an unsupervised approach to learning features.
    \item The marks are not segmented from the dataset, limiting the number of examples available, and making it difficult to train CNNs. 
\end{enumerate}

To perform visual search on the Merchant Marks, a representation for the marks that captures their structure must be learned.
\begin{figure}
    \centering
    \includegraphics[width=0.5\textwidth]{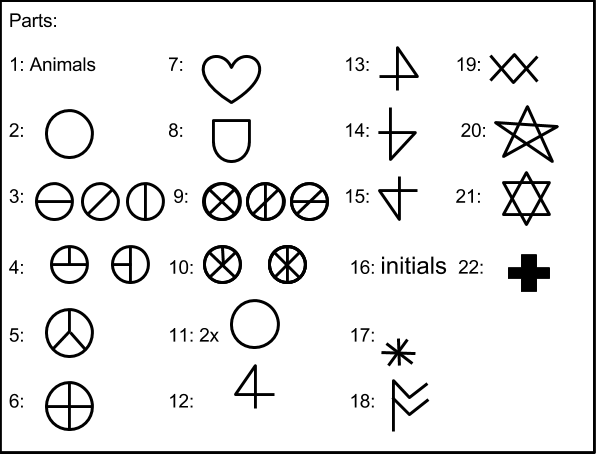}
    \caption{Most marks in the Merchant Marks dataset are made of of the above sub-structures which we refer to as \textit{parts}.}
    \label{parts}
\end{figure}
Previous work has demonstrated that deep convolutional neural networks (CNNs) are able to learn excellent hierarchical representations for data \cite{zeiler2014visualizing}. CNNs have proven useful for tasks such as classification \cite{krizhevsky2012imagenet}, segmentation \cite{noh2015learning} and have been applied to retrieval of art work \cite{Crowley14a}\cite{Crowley15}. However, these methods rely on large amounts of labelled data for learning the weights. In the absence of sufficient labelled training data, we propose the use of unsupervised techniques with CNN architectures to learn representations for the Merchant Marks.

Unlike some previous approaches in which feature representations were learned by using labelled datasets that differ in appearance from the retrieval set, we used the Merchant Marks dataset itself to learn dataset-specific features. For example, Crowley et al. \cite{Crowley14a} trained a network similar to AlexNet \cite{krizhevsky2012imagenet} on examples from the photographic scenes of ILSVRC-2012 in order to learn features for the retrieval of art work; they also trained a network on photographs of faces to learn features for retrieving paintings of faces \cite{Crowley15}.  Yu et al. \cite{yu2015sketch} suggested that features suitable for understanding natural images are not necessarily the most appropriate for understanding sketches.

Convolutional Auto-encoders (CAE) can be a useful tool for unsupervised learning of features. They are made up of two networks, an encoder which compresses the input to produce an encoding and a decoder, which reconstructs the input from that encoding. It has been shown \cite{makhzani2015winner} that shallow CAEs often learn the delta function (a trivial solution) which is not a useful representation for the data. Instead, deep encoders are needed with strict regularisation on the activations. The Winner Take All CAE \cite{makhzani2015winner} imposes both spatial and life-time sparsity on the activations of the CAE in order to learn useful representations. Other regularisation techniques include the Variational Auto-encoder \cite{kingma2013auto}, which imposes a prior distribution on the encoding.

An alternative method, which learns representations from data without the need for regularisation, is the Generative Adversarial Network \cite{goodfellow2014generative} (GAN). Deep convolutional generative adversarial networks \cite{radford2015unsupervised} have been shown to learn good representations for data. In this paper, we propose the use of GANs for learning a representation of the Merchant Marks that can be used for visual search.

The key contribution is to show that GANs can be used to learn a representation suitable for visual search.  We apply this novel idea to the Merchant Mark dataset, and compare two GAN architectures. The first GAN is designed to learn a representation for sketches, based on reported architectural considerations specific to sketches \cite{yu2015sketch}. The second GAN is a modified version of the network proposed by Radford et. al \cite{radford2015unsupervised} often used for learning representations for natural images. The representations are evaluated by comparing their invariance to shift, scale and rotation as well as the top $8$ retrieval results for $15$ examples.

\section{Generative Adversarial Networks}
\begin{figure}
    \centering
    \includegraphics[width=0.5\textwidth]{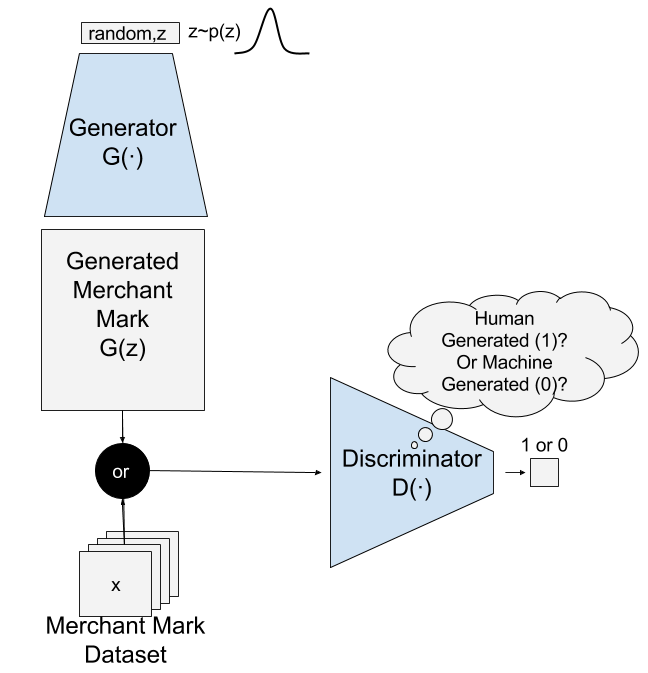}
    \caption{Generative Adversarial Network: A random sample $z$ is drawn from a prior distribution and fed into the generator, $G$ to generate a sample. The discriminator will take a sample either from the generator, $G(z)$, or from the Merchant Mark dataset, $p_{data}(x)$, and predict whether the sample is machine or human generated. The discriminator's objective is to make the correct prediction, while the generator's objective is to generate examples that fool the discriminator.}
    \label{GAN}
\end{figure}

Generative Adversarial Networks (see Fig. \ref{GAN}), (GANs) where first introduced by Goodfellow et al \cite{goodfellow2014generative}, as a generative model that learned an excellent representation for the training dataset. GANs consist of two networks, a generative network, $G$ and a discriminative network, $D$. The goal of the generative network is to learn the distribution of the training data, $p_{data}(x)$ where $x \in R^{dx}$ and $dx$ is the dimensions of a data sample.

In a GAN, the generator takes as input a vector, $z \in R^{dz}$ of $dz$ random values drawn from a prior distribution $p_z(z)$, and maps this to the data space, $G:R^{dz} \rightarrow R^{dx}$. The discriminator takes examples from both the generator and real examples of training data and predicts whether the examples are human (or real) $(1)$ or machine generated $(0)$, $D:R^{dx} \rightarrow [0,1]$. 

The objective of the discriminator is to correctly classify examples as human or machine generated, while the objective of the generator is to fool the discriminator into making incorrect predictions. This can be summarised by the following value function that the generator aims to minimise while the discriminator aims to maximise:
\[\min_{G} \max_{D} \mathbbm{E}_{x \sim p_{data(x)}} \log D(x) + \mathbbm{E}_{z \sim p_z(z)}\log (1-D(G(z))) \]

Training an adversarial network, both the generator and the discriminator learn a representation for the real data. The approach considered here will use the representation learned by the discriminator.

\section{Methods}
\begin{figure}
    \centering
    \includegraphics[width=\textwidth]{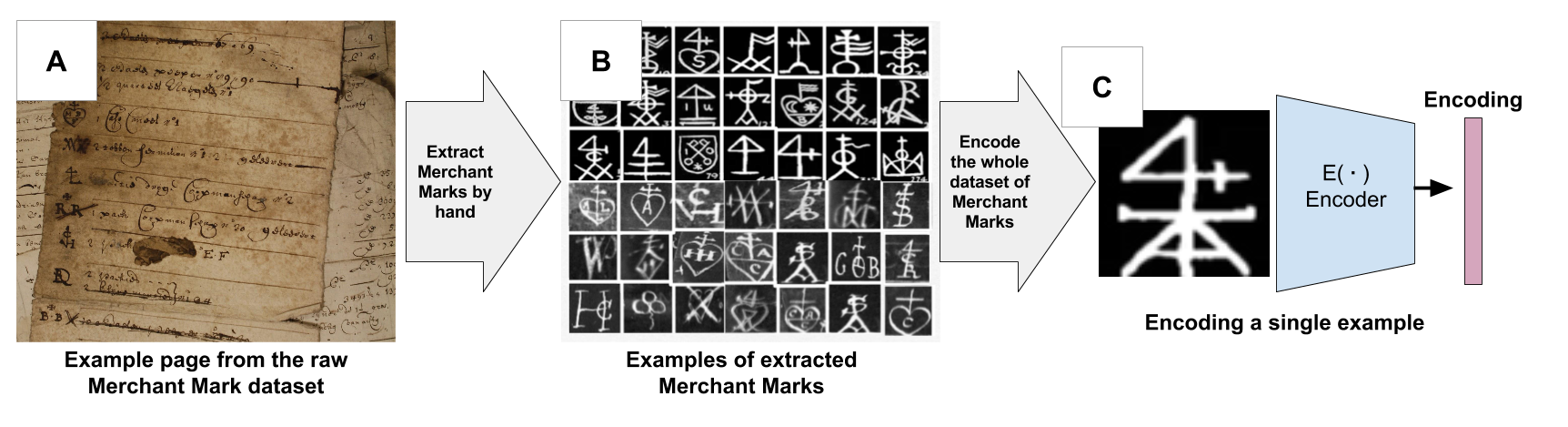}
    \caption{Overview: (A) Shows examples of raw Merchant Mark data, photographs of the documents. (B) Shows examples extracted by hand from the raw Merchant Mark dataset, a total of $2000$ examples are collected. (C) An encoder is simply a discriminator, taken from a trained GAN with the final layer removed. Representations for both query and data samples are obtained by passing examples through the encoder, the representations are used for retrieval.}
    \label{overview}
\end{figure}

Here, we show how the discriminator, taken from a trained GAN, can be modified to be used as an encoder for sketch retrieval. An overview of the methods used can be seen in Fig.\ref{overview}.

\subsection{Dataset Acquisition}
The raw Merchant Mark dataset that we have been working with consists of $76$ photographs of pages from the raw Merchant Mark dataset, similar to the one shown in Fig.\ref{overview}A, which contain multiple Merchant Marks at different spatial locations on the page. The focus of this paper is on retrieval of visually similar examples rather than localisation, so the first step involved defining box regions from which Merchant Mark training examples could be extracted. The extracted examples are re-size to be $64\times64$ pixels to form a suitable dataset for training a GAN. In total there are $2000$ training examples (see Fig.\ref{overview}B).

\subsection{Learning an Encoder}
\subsubsection{Training A GAN}
To learn an encoding, the generator and the discriminator of a GAN are trained iteratively, as proposed by Goodfellow et al. \cite{goodfellow2014generative}. See pseudo-code in Alg.1.

\begin{algorithm}
\SetAlgoLined
 \For{Number of training iterations}{
  \For{k iterations}{
  sample $p_z(z)$ to get $m$ random samples $\{z_1...z_m\}$
  
  sample $p_{data}(x)$ to get $m$ random samples $\{x_1...x_m\}$
  
  calculate the discriminator error:
  \[ J_D=-\frac{1}{2m} \left (\sum_{i=1}^m \log D(x_i) + \sum_{i=1}^m \log(1-D(G(z_i)))\right )\]
  update $\theta_D$ using Adam \cite{kingma2014adam} update rule.
  }
 sample $p_z(z)$ to get $m$ random samples $\{z_1...z_m\}$
 
 calculate the generator error:
 \[ J_G = -\frac{1}{m} \sum_{i=1}^m \log(D(G(z_i))) \]
 update $\theta_G$ using Adam \cite{kingma2014adam} update rule.
 }
 \caption{Training a GAN: After Goodfellow et al. \cite{goodfellow2014generative} with changes to the optimisation, using Adam \cite{kingma2014adam} instead of batch gradient descent. Note, $m$ is the batch size and $\theta_G$,$\theta_D$ are the weights of the generator, $G$ and discriminator, $D$.}
\end{algorithm}

\subsubsection{Network Architecture}
Both the generator, and the discriminator are convolutional neural networks \cite{springenberg2014striving}, using convolutions applied with strides rather than pooling as suggested by Radford et al. \cite{radford2015unsupervised}. 
In the discriminator, the image is mapped to a single scalar label, so the stride applied in the convolutional layer of the discriminator must be grater than $1$. A stride of $2$ is used in all convolutional layers of the discriminator. In the generator, a vector is mapped to an image, so a (positive) step size less than $1$ is needed to increase the size of the image after each convolution. A stride of $0.5$ is used in all convolutional layers of the generator.

\subsubsection{Encoding Samples} Having trained both the generator and the discriminator, the discriminator can be detached from the GAN. To encode a sample, it is passed through all but the last layer of the discriminator. The discriminative network without the final layer is called the \textit{encoder}. Both the query examples and all examples in the dataset can be encoded using the this encoder. The encoding is normalised to have unit length by dividing by the square root of the sum of squared values in the encoding.

\subsection{Retrieval}
The objective is to retrieve samples that are visually similar to a query example. To retrieve examples similar to the query, similarity measures are calculated between the representation for the query and representations for all samples in the dataset. Examples with the highest similarity scores are retrieved.
The focus of this paper is on learning a good representation for the data, for this reason a simple similarity measure is used, the (normalised) dot product.

\section{Experiments And Results}
The purpose of these experiments is to show that GANs can be used to learn a representation for our Merchant Mark dataset from only $2000$ examples, that can be used to precisely retrieve visually similar marks, given a query. 
We compare invariance of feature representations learned and retrieval results from two different networks to show that there is some benefit to using a network designed specifically for learning representations for sketches. 

\subsection{GAN Architectures}
Two different architectures were compared:
\subsubsection{sketch-GAN}
We propose a novel GAN architecture inspired by Sketch-A-Net \cite{yu2015sketch}, a network achieving state of the art recognition on sketches. Sketch-A-Net employs larger filters in the shallower layers of the discriminative network to capture structure of sketches rather than fine details which are absent in sketches. This motivated our network design, using larger filters in the lower levels of the discriminator and the deeper levels of the generator. This network will be referred to as the \textit{sketch-GAN}. This network has only $33$k parameters.

\subsubsection{thin-GAN} A network similar to that proposed by Radford et al. \cite{radford2015unsupervised} is used. This network has very small filters, consistent with most of the state-of-the-art natural image recognition networks \cite{simonyan2014very}. The original network has 12.4M parameters which would not compare fairly with the \textit{sketch-GAN}, instead a network with $1/16$th of the filters in each layer is used, this will be referred to as the \textit{thin-GAN} and has $50$k parameters. Full details of the architecture are given in Table.\ref{arch}.

\setlength{\tabcolsep}{4pt}
\begin{table}
\begin{center}
\caption{A summary of the network architectures used in this study. fc=fully connected layer, c=convolutional layer with stride 2, d=convolutional layer with stride 0.5, unless stated otherwise; for all cases, $dz$, the dimension of the random valued vector input to the generator is 2. The ReLU activation function is used in all hidden layers of all networks and the sigmoid activation is used in final layer of each network.}
\label{arch}
\begin{tabular}{llll}
\hline\noalign{\smallskip}
thin-GAN:G & thin-GAN:D\\
\noalign{\smallskip}
\hline
\noalign{\smallskip}
fc: $1024 \times dz $, reshape(64,4,4)     & c: $8 \times 1 \times 3 \times 3$  \\   
d: $32 \times 64 \times 3 \times 3$  & c: $16 \times 8 \times 3 \times 3$ \\
batch normalisation                     & batch normalisation                   \\  
d: $16 \times 32 \times 3 \times 3$   & c: $32 \times 16 \times 3 \times 3$ \\
batch normalisation                     & batch normalisation                   \\
d: $8 \times 16 \times 3 \times 3$   & c: $64 \times 32 \times 3 \times 3$\\
batch normalisation                     & batch normalisation, reshape(1024)          \\
d: $1 \times 8 \times 3 \times 3$     & fc: $1 \times 1024$               \\

\hline\noalign{\smallskip}
sketch-GAN:G & sketch-GAN:D\\
\noalign{\smallskip}
\hline
\noalign{\smallskip}
fc: $128 \times dz$, reshape(8,4,4)     & c: $8 \times 1 \times 9 \times 9$ (stride=1)\\
d: $16 \times 8 \times 3 \times 3$  & c: $16 \times 8 \times 5 \times 5$\\
batch normalisation                 & batch normalisation\\
d: $16 \times 16 \times 5 \times 5$ & c: $16 \times 16 \times 5 \times 5$\\
batch normalisation                 & batch normalisation\\
d: $16 \times 16 \times 5 \times 5$ & c: $16 \times 16 \times 5 \times 5$\\
batch normalisation                 & batch normalisation, reshape(1,1024)\\
d: $16 \times 16 \times 5 \times 5$ & fc: $1 \times 1024$ \\
batch normalisation                 & - \\
d: $1 \times 16 \times 9 \times 9$ (stride=1)  & - \\

\hline
\end{tabular}
\end{center}
\end{table}
\setlength{\tabcolsep}{1.4pt}

\subsection{Details of Training}
In adversarial training the generator and discriminator networks are competing against eachother in a mini-max game, where the optimal solution is a Nash Equilibrium \cite{salimans2016improved}. Adversarial networks are trained iteratively alternating between the generator and discriminator using gradient descent which aims to minimise the individual cost functions of the generator and discriminator, rather than finding a Nash Equilibrium \cite{salimans2016improved}. For this reason convergence, during adversarial training cannot be guaranteed \cite{salimans2016improved}\cite{goodfellow2014distinguishability}. During training we found that networks did not converge, for this reason networks were trained for a fixed number of iterations, rather than till the networks converged. The networks are trained for $2000$ iterations with batch size of $128$ according to Alg. 1 \cite{goodfellow2014generative}, with $k=1$, $dz=2$, learning rate = $0.002$, and $p_z(z) \sim U(0,1)$. The networks were still able to learn features useful for retrieval despite not converging.

\subsection{Feature Invariance}
Merchant Marks are hand drawn, which means that the marks are likely to vary in both scale and orientation. It is therefore important to consider the rotation and scale invariance of the representations that result from training. When searching a document for Marks, one approach may be to apply a sliding box search. The step size in sliding the search box will affect the computational feasibility of the search. If a representation used for search is invariant to larger shifts, then a sliding box search can be performed with a larger step size, making the search more efficient. For this reason, shift invariance of the two representations is also compared.

\begin{figure}
    \centering
    \includegraphics[width=\textwidth]{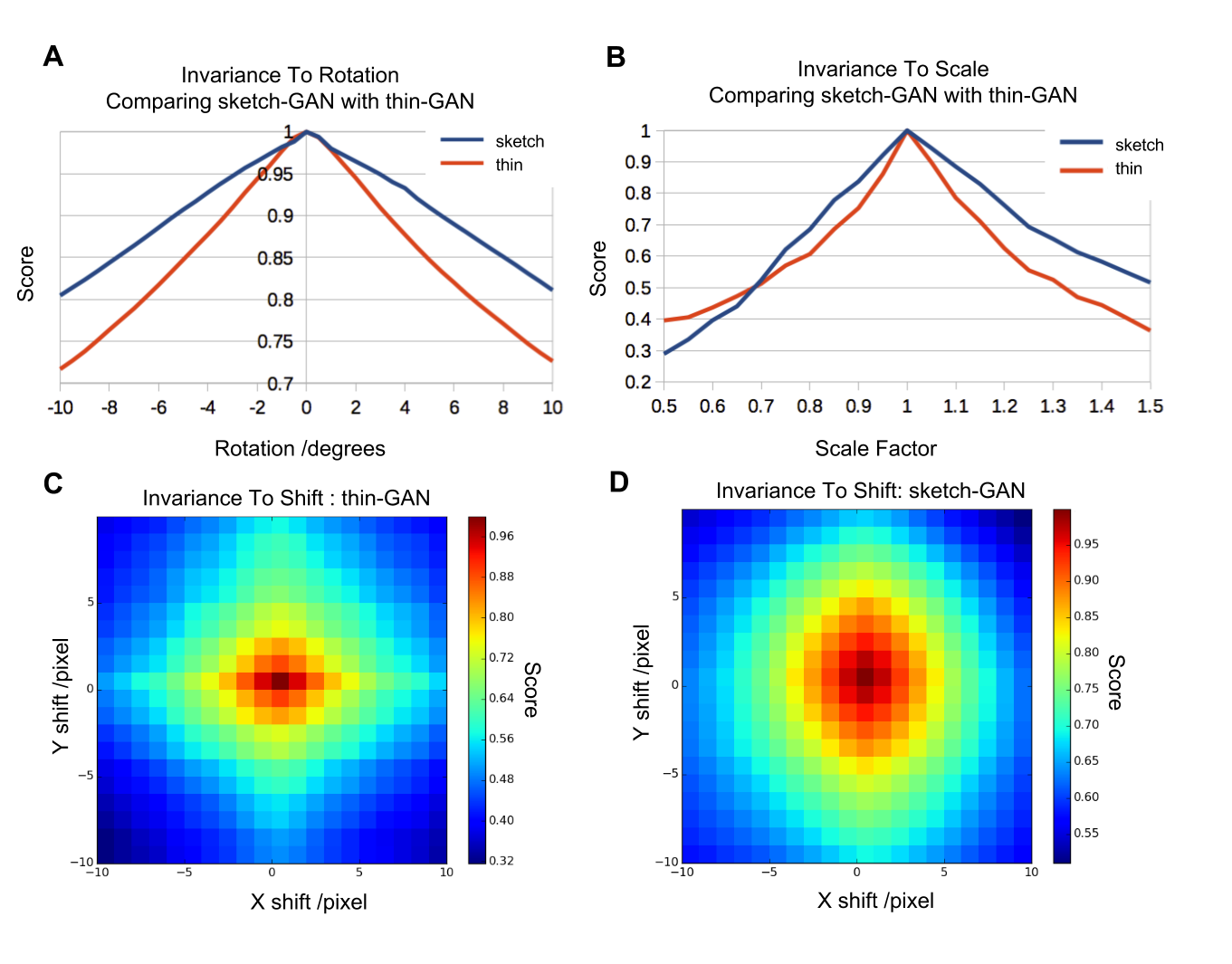}
    \caption{Invariance: Shows invariance of the \textit{sketch-GAN} and \textit{thin-GAN} representations to A) rotation, B) scale and C,D) translation.} 
    \label{invariances}
\end{figure}

\subsubsection{Invariance To Rotation}
To assess the degree of rotation invariance within the two representations, $100$ samples were randomly taken from the Merchant Mark dataset and rotated between the angles of $-10$ and $10$ degrees. At each $0.5$ degree increment, the samples were encoded and the similarity score between the rotated sample and the sample at $0$ degrees was calculated. The similarity score used was the normalised dot product, since this was also the measure used for retrieval. The results are shown in the top left of Fig.\ref{invariances}. It is clear that the \textit{sketch-GAN} encoding is more tolerant to rotation than the \textit{thin-GAN} encoding. Note that the background of the rotated samples were set to $0$ to match the background of the samples.

\subsubsection{Invariance To Scale}
A similar approach was used to assess the degree of scale invariance within the two networks. Again, $100$ samples were randomly taken from the Merchant Mark dataset, and scaled by a factor between $0.5$ and $1.5$. At each increment of $0.05$, the scaled sample was encoded and a similarity score was calculated between the scaled samples and the sample at scale $1$. The results are shown in the top right of Fig.\ref{invariances}. Note, that when the scaling factor is $<1$ the scaled image is padded with zeros to preserve the $64 \times 64$ image size. When scaling with a factor $>1$, the example is scaled and cropped to be of size $64 \times 64$. The bounding box of the unscaled marks is tight, which means that at higher scaling factors parts of the marks are sometimes cropped out. Despite this, the \textit{sketch-GAN} encoding is able to cope better with up-scaling compared to down-scaling. The \textit{sketch-GAN} encoder generally outperforms the \textit{thin-GAN} encoder, particularly for up-scaling.

\subsubsection{Invariance To Shift}
Finally, we compared the shift invariance of the two encoders. Sampling $100$ marks from the merchant mark dataset, and applying shifts between $-10$ and $10$ pixels in increments of $1$ pixel in both the $x$ and $y$ directions. The results are shown as a heat map in Fig.\ref{invariances}, where the \textit{sketch-GAN} encoding appears to be more invariant to shift than the \textit{thin-GAN} encoding. 

\subsection{Retrieval}
For the retrieval experiments, $500$ queries were taken at random from the training dataset and used to query the whole dataset using features from the \textit{sketch-GAN}. The top $9$ matches were retrieved, where the first retrieval is the example itself and the rest are examples that the system thinks are similar. The results from some of these queries are shown in Fig.\ref{sketchGANret}.
The same query examples were used to query the dataset using the features from the \textit{thin-GAN}, the results of these queries are shown in Fig.\ref{GANret}.

\subsubsection{Retrieval results using trained sketch-GAN encoder} 
\begin{figure}[!tbp]
  \centering
  \subfloat[thin-GAN]{\includegraphics[width=0.45\textwidth]{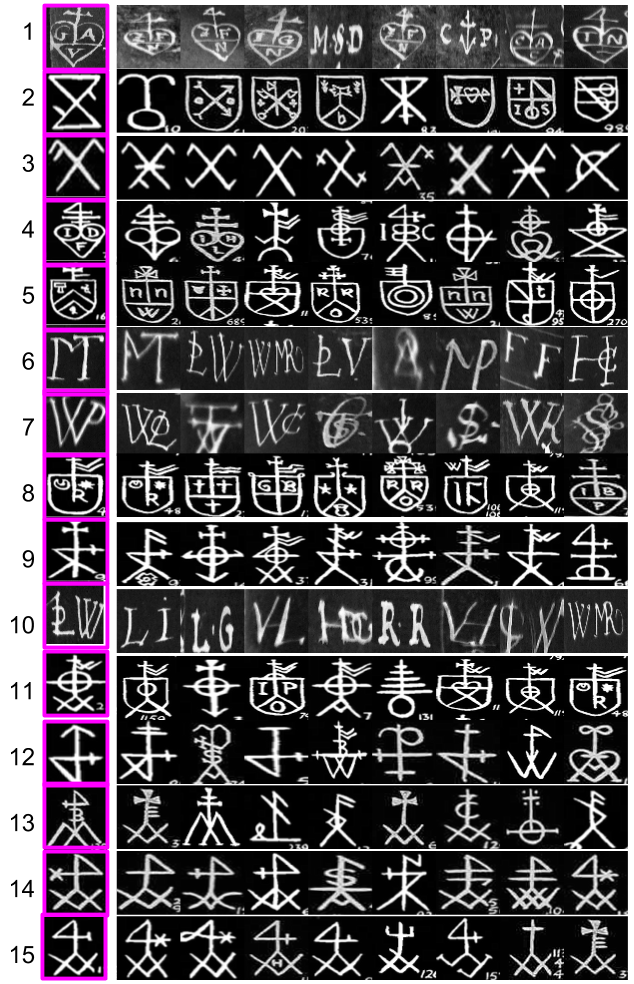}\label{GANret}}
  \hfill
  \subfloat[sketch-GAN]{\includegraphics[width=0.45\textwidth]{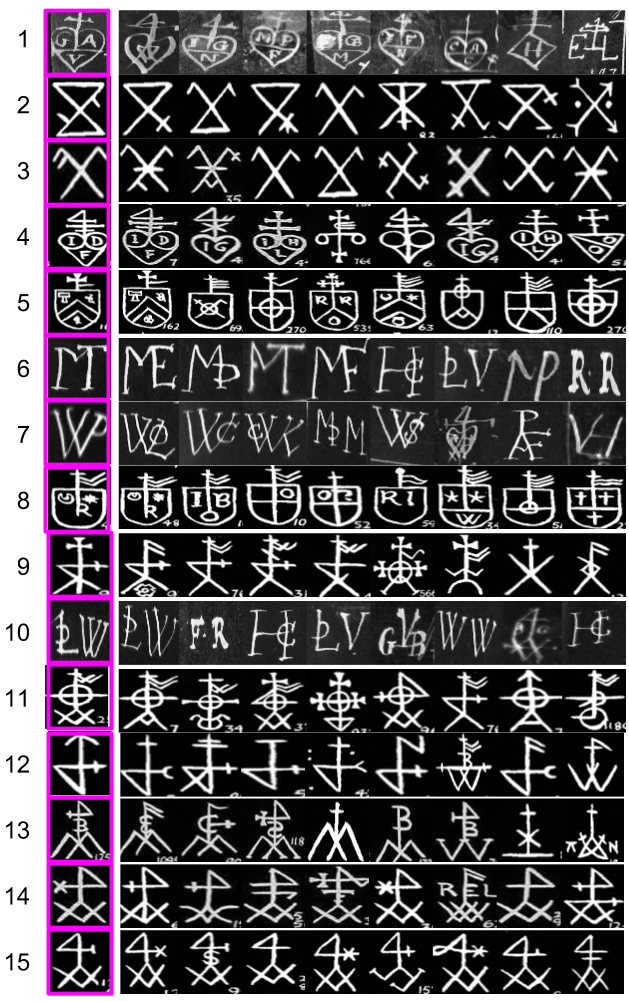}\label{sketchGANret}}
  \caption{Retrieval examples using different GAN architectures. Each sub-figure shows 15 retrievals where the 1st example, in a purple box, in each row is the query and the following images on each row are the top 8 retrievals. (a) Shows retrievals using the thin-GAN encoder and (b) shows retrievals using the sketch-GAN encoder.}
\end{figure}

Results show that using the \textit{sketch-GAN} encoder for Merchant Marks retrieval (Fig.\ref{sketchGANret}) allows retrieval of examples that have multiple similar parts for example results for queries $\#4$,$\#8$,$\#11$,$\#14$ and $\#15$ consistently retrieve examples with at least two similar parts (Fig.\ref{parts}). Specifically, most retrievals for query $\#15$, Fig.\ref{sketchGANret} have parts $12$ and $19$ from Fig.\ref{parts}. Exact matches are found for retrievals $\#4$,$\#5$,$\#8$ and $\#10$. Specifically, query $\#10$ finds an exact match despite the most similar example being shifted upwards and rotated slightly. Retrievals for query \#$6$ finds an exact match but does not rank the retrieval as high as non-exact matches, suggesting that there is still room for improvement in the representations that are learned.

\subsubsection{Retrieval results using trained thin-GAN encoder} 
On visual inspection of the retrieval results that use the \textit{thin-GAN} encoder, it is clear that they under perform compared to the \textit{sketch-GAN} for the same query examples, with fewer visually similar examples. The \textit{thin-GAN} encoder fails to find exact matches for $4$,$5$ and $10$. Failure to find a match for $10$ further suggests that the \textit{thin-GAN} is less invariant to rotation.

\section{Conclusions}
Convolutional networks contain, at a minimum, tens of thousands of weights. Training such networks has typically relied on the availability of large quantities of labelled data.  Learning network weights that provide good image representations in the absence of class labels is an attractive proposition for many problems. One approach to training in the absence of class labels is to encourage networks to compete in coupled tasks of image synthesis and discrimination.  The question is whether such Generative Adversarial Networks can learn feature representations suitable for retrieval in a way that matches human perception.\\

We have found that GANs can indeed be used to learn representations that are suitable for image retrieval. To demonstrate this, we compared the representation learned by GANs that were trained on Merchant Marks. We compared two related architectures, \textit{sketch-GAN} and \textit{thin-GAN}; \textit{sketch-GAN} has an architectural design that is more appropriate for performing generation and discrimination of sketches. Our experiments showed that GANs are suitable for retrieval of both visually similar and exact examples. Experiments also showed that the features that were learned by the \textit{sketch-GAN} were, on average, more robust to small image perturbations in scale, rotation and shift than the \textit{thin-GAN}. Further, retrieval results when using the \textit{sketch-GAN} appeared more consistent than in using \textit{thin-GAN}.\\

More generally, the experiments suggest that adversarial training can be used to train convolutional networks for the purpose of learning good representations for the retrieval of perceptually similar samples; this can be achieved without  the level of labelling and examples required for non-adversarial training approaches.  This broadens the scope of deep networks to problems of perceptually similar retrieval in the absence of class labels, a problem that is increasingly of interest in heritage collections of images.

\subsubsection*{Acknowledgements}
We would like to acknowledge Nancy Bell, Head of Collection Care at the National Archives. We would also like to acknowledge the Engineering and Physical Sciences Research Council for funding through a Doctoral Training studentship.


\bibliographystyle{splncs03}
\bibliography{eccv2016submission}
\end{document}